\pdfoutput=1

\documentclass[11pt]{article}

\usepackage[preprint]{acl}
\usepackage{booktabs}
\usepackage{algorithm}
\usepackage{tikz}
\usepackage{pgfplots}
\usepackage{algpseudocode}
\usepackage{times}
\usepackage{latexsym}
\usepackage{multirow}
\usepackage{amsmath}
\usepackage[T1]{fontenc}

\usepackage[utf8]{inputenc}

\usepackage{microtype}

\usepackage{inconsolata}

\usepackage{graphicx}
\usepackage{breakurl}
\usepackage{xcolor}
\usepackage{colortbl}
\usepackage{multirow}
\usepackage{graphicx}
\usepackage{tabularx}
\usepackage{amssymb}
\usepackage{hyperref}

\makeatletter
\title{%
  \vspace{-2cm}
  \begin{center}
    \includegraphics[height=2cm]{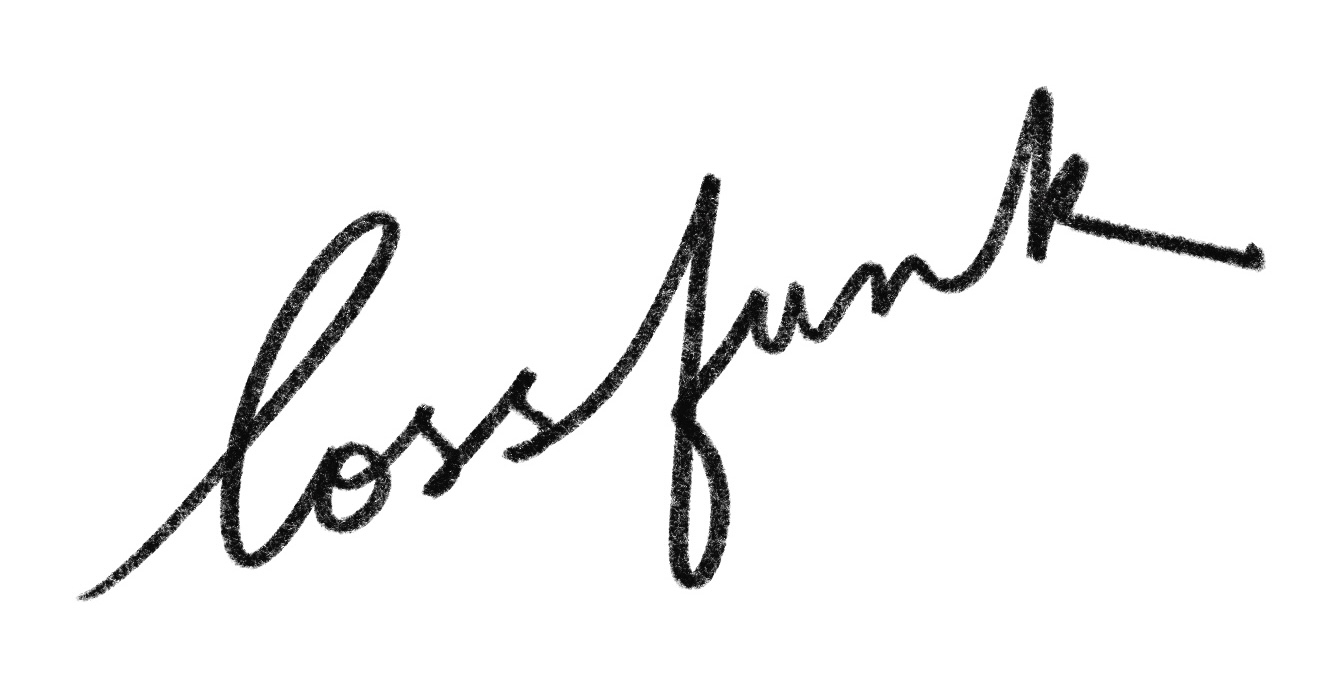}
  \end{center}
  \vspace{0.5cm}
  IPO: Your Language Model is Secretly a Preference Classifier%
}
\makeatother

\author{
  \textbf{Shivank Garg\textsuperscript{1,2}\textsuperscript{*}},
  \textbf{Ayush Singh\textsuperscript{1,2}\textsuperscript{*}},
  \textbf{Shweta Singh\textsuperscript{1,2}},
  \textbf{Paras Chopra\textsuperscript{2}}
\\
  \textsuperscript{1}Indian Institute of Technology Roorkee,
  \textsuperscript{2}Lossfunk
\\
  \texttt{
    \texttt{\{shivank\_g@mfs, ayush\_s@mt, shweta\_s@mfs\}.iitr.ac.in}},
    \texttt{paras@lossfunk.com}
    \\
  }

\begin{document}

\maketitle
\begin{abstract}

\footnotetext{*Equal contribution.}

Reinforcement learning from human feedback (RLHF) has emerged as the primary method for aligning large language models (LLMs) with human preferences.
While it enables LLMs to achieve human-level alignment, it often incurs significant computational and financial costs due to its reliance on training external reward models or human-labeled preferences. In this work, we propose \textbf{Implicit Preference Optimization (IPO)}, an alternative approach that leverages generative LLMs as preference classifiers, thereby reducing the dependence on external human feedback or reward models to obtain preferences. We conduct a comprehensive evaluation on the preference classification ability of LLMs using RewardBench, assessing models across different sizes, architectures, and training levels to validate our hypothesis. Furthermore, we investigate the self-improvement capabilities of LLMs by generating multiple responses for a given instruction and employing the model itself as a preference classifier for Direct Preference Optimization (DPO)-based training. Our findings demonstrate that models trained through IPO achieve performance comparable to those utilizing state-of-the-art reward models for obtaining preferences. Our code is available at \textcolor{magenta}{\burl{https://github.com/shivank21/Implicit\_Preference\_Optimization}}.
\end{abstract}

\section{Introduction}

Large Language Models (LLMs) such as GPT4 \citep{openai2024gpt4technicalreport}, Gemini \citep{geminiteam2024gemini15unlockingmultimodal}, and Llama \citep{touvron2023llama2openfoundation} have become highly popular due to their remarkable capabilities. These models often rely on two key techniques: Reinforcement Learning from Human Feedback (RLHF) and Inference Scaling. Reward models are central to both approaches. In RLHF, reward models act as proxies for human values, providing feedback on generated text to align language models during training \cite{christiano2023deepreinforcementlearninghuman,ziegler2020finetuninglanguagemodelshuman}. Similarly, in inference scaling, reward models are used to select the best response from a set of candidates based on predicted rewards \cite{snell2024scalingllmtesttimecompute}.

The training of reward models, however, relies heavily on high-quality, human-generated data, which is both costly and time-intensive. To address this limitation, recent works have explored Reinforcement Learning from AI Feedback (RLAIF) \citep{lee2023rlaif}, where AI-generated feedback is used to train reward models. This approach reduces the dependency on human-annotated data but introduces challenges, including heuristic assumptions that LLMs can consistently provide high-quality feedback and the requirement for larger LLMs to generate such feedback \citep{pang2023language}.

\begin{figure*}[h!]
    \centering
 \includegraphics[width=\textwidth]{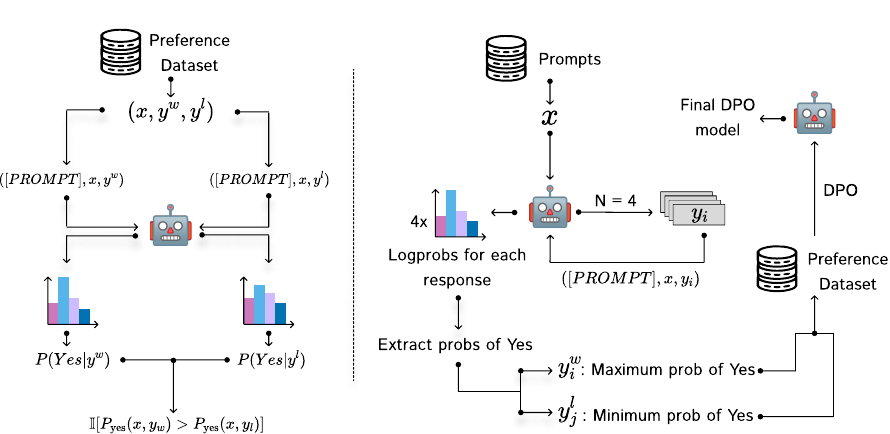}
 \caption{\textbf{Left}: We evaluate preferences using \textit{(Prompt, Chosen, Rejected)} triplets, scoring responses based on the probability of the token "Yes" given classification prompt. The evaluation is correct if the Chosen response scores higher than the Rejected oner. Here [PROMPT] refers to the category specific prompt. \textbf{Right}: Our Self-Improving DPO framework generates diverse responses, rates them, constructs a preference dataset, and trains the model via DPO.}
    \label{Approach}
\end{figure*}

Self-rewarding large language models  \citep{yuan2024selfrewarding} have emerged as a promising alternative for improving language model performance. In this paradigm, a single model assumes dual roles: as an actor, it generates responses to fulfill specific instructions, and as a judge, it evaluates these responses using the LLM-as-a-Judge framework \citep{zheng2023judging} to assign rewards. However, despite its potential, this approach has a fundamental limitation—the model undergoes fine-tuning to improve its response generation but not its evaluative capabilities. As a result, while it evolves as an actor, its ability to judge remains static.

To address this limitation, Meta-Rewarding Language Models \citep{wu2024meta} extend the model's judging capabilities by explicitly fine-tuning it for judging responses. Additionally, approaches such as Self-Evolving Reward Models \citep{huang2024self} introduce a data-filtering pipeline that leverages high-quality model-generated outputs to refine reward model training. Nevertheless, a significant challenge with these methods lies in their dependence on discrete reward signals or the necessity of external models and datasets, which may introduce inefficiencies or constraints in scalability.

We hypothesize that providing a preference magnitude, rather than discrete prompt based feedback, enables more fine-grained evaluation of model responses. Drawing inspiration from VQA score \citep{lin2025evaluating}, we introduce a probabilistic framework for rewarding LLM-generated responses. This framework empowers even base models to assess and assign rewards to responses, effectively allowing them to function as preference classifiers without relying on external reward models. Compared to existing prompting-based preference strategies, which require large LLMs to act as judges through explicit prompting, our approach is more computationally efficient. It eliminates the need for external supervision or additional training. Specifically, we propose \textit{Implicit Preference Optimization (IPO)}, a novel framework that demonstrates how any LLM can serve as an effective preference classifier.

We conduct extensive experiments across multiple model families, including Qwen, LLaMA, Mistral, and GPT, encompassing various model sizes and configurations (base and instruction-tuned). Additionally, we evaluate our approach on math and code-specific models to analyze their effectiveness as preference classifiers. To rigorously assess our hypothesis of LLM as a preference classifier, we benchmark the ability of LLM to model preferences using RewardBench, a standardized reward model evaluation suite. \textbf{Our findings indicate that LLMs can perform well as preference classifiers, achieving accuracy levels surpassing those of several reward models \citep{lambert2024rewardbenchevaluatingrewardmodels}.} 

Moreover, previous work has highlighted the challenges of training efficient reward models for code and maths-related tasks. Our findings suggest that both general-purpose and code-specific models can inherently function as effective preference classifiers; however, math-specific models lack this ability. To further validate this hypothesis, we examine IPO within a self-improving model setup, where the model generates responses, ranks them based on its own preferences, and leverages these rankings for Direct Preference Optimization (DPO)-based training. Our results demonstrate the effectiveness of IPO in improving response quality.

\section{Background and Related Work}

\subsection{Reinforcement Learning for Improving LLMs}

Recent approaches for improving LLMs involve training a fixed reward model using human preference data, which is subsequently utilized for Reinforcement Learning (RL) to train language models. This method, commonly referred to as Reinforcement Learning from Human Feedback (RLHF) \citep{liu2020learning,ouyang2022training}, has significantly enhanced the performance of models like Llama\citep{touvron2023llama2openfoundation,dubey2024llama} and ChatGPT\citep{openai2024gpt4technicalreport}.

An alternative paradigm to traditional RLHF are methods like Direct Preference Optimization (DPO) \cite{rafailov2024directpreferenceoptimizationlanguage}, which bypasses the need for training a reward model altogether. Instead, it directly trains the LLM based on human preference data. Beyond RLHF and DPO, additional techniques such as Kahneman \& Tversky’s Optimization (KTO) \citep{ethayarajh2024ktomodelalignmentprospect}, Sequence Likelihood Calibration (SLiC) \citep{zhao2023slichfsequencelikelihoodcalibration}, Reinforced Self-Training (ReST) \citep{gulcehre2023reinforcedselftrainingrestlanguage}, and Rank Responses with Human Feedback (RRHF) \citep{yuan2023rrhfrankresponsesalign} have been proposed, each leveraging human preferences to optimize LLM training.

Constitutional AI \citep{bai2022constitutional} uses an LLM to provide feedback to refine responses. The feedback is then used to further train the language model through Reinforcement Learning from AI Feedback (RLAIF) \citep{lee2023rlaif}. Similarly, Self-Play fIne-tuNing (SPIN) \citep{chen2024selfplayfinetuningconvertsweak} introduces an Interactive DPO-like framework, designed to eliminate the need for reward model training and to simplify reliance on human-labeled data pairs.

\subsection{Self Improving Models}

Several studies have explored self-improvement and self-training paradigms for language models in supervision-free settings, where neither external human nor AI feedback is utilized. Works such as LMSI \citep{huang2022largelanguagemodelsselfimprove, huang2024selfimprovementlanguagemodelssharpening} investigate techniques that enable language models to autonomously enhance their own performance without relying on explicit annotations or reward signals.

The concept of \textit{LLM-as-a-Judge} \citep{gu2024surveyllmasajudge,ye2024beyond,dong-etal-2024-llm,li2024dissecting} has also been extensively studied, where various methods have been proposed to design self-rewarding reward functions, denoted as $r_{self}$, using carefully crafted prompting strategies. These approaches aim to enable language models to evaluate their own outputs effectively, thereby facilitating self-refinement.

In addition to these works, ResT-MCTS\textsuperscript{*} \citep{zhang2024rest} and SPPO \citep{wu2024self} have explored algorithms based on self-training and self-play, where models iteratively improve their own performance through interaction with generated data. While these methods emphasize self-guidance, many incorporate external feedback mechanisms, such as Supervised Fine-Tuning (SFT) or reward-based optimization, to further refine the training process \cite{ouyang2022training}. 

\subsection{Evaluation of Reward Models}

Evaluating reward models plays a crucial role in aligning large language models (LLMs) with human preferences. Various works, such as AlpacaFarm \citep{dubois2024alpacafarmsimulationframeworkmethods}, evaluate preference models by comparing model-generated outputs with those from a reference model. Similarly, ChatbotArena \citep{chiang2024chatbotarenaopenplatform} determines preferences between two model-generated outputs. These methods, however, focus on indirectly evaluating reward models rather than conducting direct evaluations.

Recent benchmarks, such as RewardBench \citep{lambert2024rewardbenchevaluatingrewardmodels} and RM-Bench \citep{liu2024rmbenchbenchmarkingrewardmodels}, address this gap by creating category-wise, high-quality binary datasets to model and evaluate reward model performance. Given the robustness and high quality of these datasets, we use them to test our hypothesis.




\section{LLM as Preference Model}
\label{LLM_as_Preference}
\subsection{Background}

Large Language Models (LLMs) generate text in an autoregressive manner, producing tokens sequentially based on the context of previously generated tokens. Given an input context \( \mathbf{x} \) , the autoregressive model predicts an output sequence \( \mathbf{y} = (y_1, y_2, \dots, y_T) \) one token at a time. Assuming the model is parameterized by \( \theta \), the conditional probability of generating the sequence \( \mathbf{y} \) is defined as:

\begin{equation}
    p_\theta(\mathbf{y} \mid \mathbf{x}) = \prod_{t=1}^T p_\theta(y_t \mid \mathbf{x}, y_{<t}),
\end{equation}

where \( y_{<t} = (y_1, y_2, \dots, y_{t-1}) \). For notational simplicity, \( p_\theta(y_t \mid \mathbf{x}) \) is used to represent \( p_\theta(y_t \mid \mathbf{x}, y_{<t}) \).

The probability distribution over the vocabulary at each time step \( t \) is computed using a softmax function on the logits \( z \) as:

\begin{equation}
    p_\theta(y_t \mid \mathbf{x}) = \frac{\exp(z_t / \tau)}{\sum_{i=1}^M \exp(z_i / \tau)},
\end{equation}

where \( z_t = \text{logit}_\theta(y_t \mid \mathbf{x}, y_{<t}) \), \( M \) is the vocabulary size, and \( \tau > 0 \) is a temperature parameter. 

Various decoding strategies govern token selection during text generation. Greedy decoding selects the highest probability token at each step, while beam search expands multiple candidate sequences in parallel to find the most likely one. Top-k sampling \citep{fan2018hierarchical}, on the other hand, limits token choices to the k most probable candidates, introducing diversity. Many other decoding strategies also exist, each balancing fluency and variability differently.

\begin{table*}[htbp!]
\centering
\setlength{\tabcolsep}{4pt} 
\definecolor{apricot}{rgb}{0.95, 0.82, 0.62}
\definecolor{lightgray}{rgb}{0.96, 0.96, 0.96}

\begin{tabular}{l | ccccc | ccccc }
\hline
\rowcolor{apricot} 
\multirow{0}{*}{\textbf{Models}} & \multicolumn{5}{c|}{\textbf{Our Approach}} & \multicolumn{5}{c}{\textbf{Self Rewarding}} \\
\cmidrule(lr){2-6} \cmidrule(lr){7-11}
 & Chat & Code & Math & Safety & Average & Chat & Code & Math & Safety & Average  \\ \hline
Llama-3.2-1B-Inst & 64.37 & 52.84 & 88.14 & 80.48 & 71.45 & 30.47 & 21.03 & 14.54 & 31.55 & 24.39\\ \hline
\rowcolor{lightgray}Llama-3.2-3B-Inst & 62.09 & 67.17 & \textbf{98.21} & 80.23 & 76.92 & 33.87 & 24.69 & 36.01 & 46.73 & 35.32\\ \hline
Llama-3-8B-Inst & 59.56 & 73.88 & 54.97 & 87.88 & 69.07 & 35.43 & 12.29 & 21.70 & \textbf{58.35} & 31.94\\ \hline
\rowcolor{lightgray}Qwen-2.5-3B-Inst & 60.89  & 80.59  & 46.31  & 86.05 & 68.46 & 26.72 & 23.88 & \textbf{41.61} & 24.43 & 29.16\\ \hline
Qwen-2.5-7B-Inst & \textbf{78.26}  & \textbf{83.13}  & 56.24  & \textbf{93.24} & 77.71 & \textbf{58.73} & \textbf{47.93} & 40.49 & 52.20 & \textbf{49.82}\\ \hline
\rowcolor{lightgray}Mistral-7B-Inst & 61.25 & 70.93  & 96.20 & 83.85 & \textbf{78.05} & 24.55 & 1.6 & 28.18 & 15.39 &  17.43\\ \hline
Gemma2-2B-It & 35.34 & 42.58  & 91.50 & 70.04 & 59.86 & 22.36 & 2.84 & 12.75 & 34.78 & 18.18\\ \hline
\rowcolor{lightgray}Phi-3-Mini-Instruct  & 55.91 & 75.30 & 89.10 & 75.32 & 73.90 & 46.63 & 35.46 & 22.60 & 56.75 & 40.36 \\ \hline
\end{tabular}
\caption{The above table compares our approach with the Self Rewarding approach. The row labels correspond to the model name and the column labels correspond to the sub-categories. The metric used is accuracy where the higher values indicate better performance. }
\label{tab:reward_bench}
\end{table*}

\subsection{Methodology}
Our approach leverages a language model as a preference model, evaluating response appropriateness through binary classification. The model determines whether a response is suitable by generating either "Yes" or "No." To guide this assessment, we employ category-specific prompts, which are detailed in Appendix \ref{Prompts_table}. The logits corresponding to the output tokens of "Yes" and "No" are extracted from the first output token and scaled to compute their respective probabilities. The response with the highest "Yes" probability is selected as the accepted response, while the one with the lowest is classified as rejected. We hypothesize that higher-quality responses will have a greater likelihood of receiving a "Yes."




\begin{figure}[h]
    \centering
    \includegraphics[width=\columnwidth]{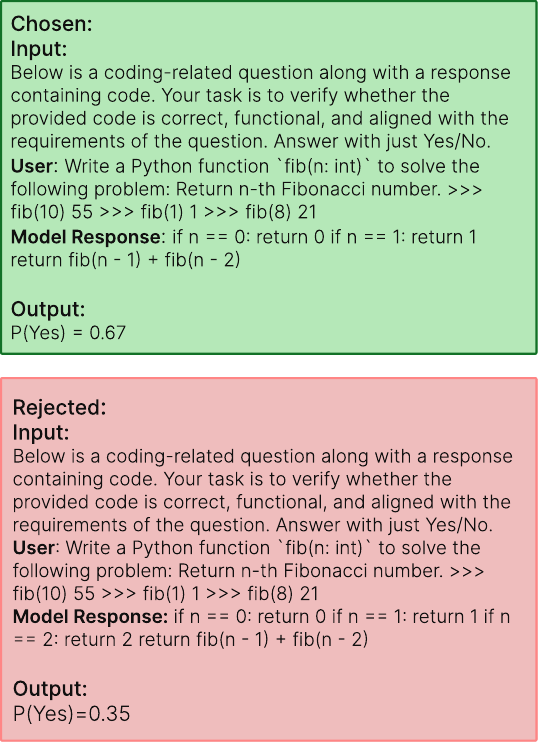}
    \caption{Example outputs from Reward Bench using our approach.}
    \label{Sample_Prompts}
\end{figure}

\subsubsection{Preference Classification}
\label{Preference Modeling}

In our experiments, we observed that guiding the language model to initiate its responses with "Yes" or "No" was essential, particularly for smaller models whose outputs are highly sensitive to prompt phrasing. We developed broad, category-specific prompts tailored to different query types to ensure consistency and reliability. Following prior research \citep{lambert2024rewardbenchevaluatingrewardmodels, liu2024rmbenchbenchmarkingrewardmodels}, we classify prompts into four overarching categories: Code, Math, Chat and Safety. Additional details about prompts are provided in Appendix \ref{Prompts_table}. An example prompt is shown in Figure \ref{Sample_Prompts}.

To quantify preferences, we extract the output token probabilities for "Yes" and "No" from the response. The detailed approach is outlined below:

Given an input token sequence \( \mathbf{x} = (x_1, x_2, \dots, x_T) \), a language model \( f(\cdot) \) generates a probability distribution over the vocabulary \( \mathcal{V} \) for the next token. Specifically, the model outputs a logit vector \( \mathbf{z} \in \mathbb{R}^{|\mathcal{V}|} \), where  

\begin{equation}
\mathbf{z} = f(\mathbf{x}).
\end{equation}

To derive probabilities, we apply the softmax function over the logits:  

\begin{equation}
p_i = \frac{\exp(z_i)}{\sum_{j \in \mathcal{V}} \exp(z_j)}, \quad \forall i \in \mathcal{V},
\end{equation}

where \( p_i \) represents the probability assigned to token \( i \). Thus we define probability of "Yes" token as $p_{\text{yes}}$ and "No" token as $p_{\text{no}}$. Then we normalize the probabilities to ensure a fair comparison:



\begin{equation}
p_{\text{yes}}' = \frac{p_{\text{yes}}}{p_{\text{yes}} + p_{\text{no}}}, \quad 
p_{\text{no}}' = \frac{p_{\text{no}}}{p_{\text{yes}} + p_{\text{no}}}.
\end{equation}

The final values \( (p_{\text{yes}}', p_{\text{no}}') \) represent the normalized likelihoods of the model predicting "Yes" or "No" .

\subsection{Experiments}
\subsubsection{Benchmarking Our Approach}

To evaluate our approach, we conducted experiments using LLMs of varying sizes and architectures. We compared instruction-tuned models with their base counterparts. Additionally, we analyzed the effect of fine-tuning on a specialized task like code/math problems on preference classification by including models fine-tuned for these tasks. For comparisons involving a reward model we use the Skywork Reward Llama 8B model \cite{liu2024skywork} as the baseline.
The detailed results for all the comparisons are available in Appendix \ref{All Results}.

In particular, we tested the following models:

\begin{itemize}
  \setlength{\itemsep}{0.05em} 
  \item \textbf{LLaMA Family \citep{dubey2024llama}:} LLaMA-3.2-1B, LLaMA-3.2-1B-Instruct, LLaMA-3.2-3B, LLaMA-3.2-3B-Instruct, Meta LLaMA 3-8B, Meta LLaMA 3-8B-Instruct.
  \item \textbf{Mistral Family \citep{jiang2023mistral}:} Mistral 7B, Mistral 7B-Instruct.
  \item \textbf{Qwen Family \citep{yang2024qwen2}:} Qwen2.5-3B, Qwen2.5-3B-Instruct, Qwen2.5-7B, Qwen2.5-7B-Instruct.
  \item \textbf{Code Generation Models:} Starcoder2-7B \citep{lozhkov2024starcoder}, CodeGemma-7B-It \citep{team2024codegemma}, Qwen-Coder-7B-Inst \citep{hui2024qwen2}, Qwen-Coder-3B-Inst.
  \item \textbf{Math Generation Models:} Qwen-Math-7B-Inst, Qwen-Math-1.5B-Instruct \citep{yang2024qwen2}, Deepseek-Math-7B \citep{shao2024deepseekmathpushinglimitsmathematical}, Llemma-7B \citep{azerbayev2024llemmaopenlanguagemodel}.
  \item \textbf{Other Models:} Phi-3-mini-128k-Instruct \citep{abdin2024phi}, Gemma 2B-Instruct \citep{team2024gemma}, GPT-4o Mini \citep{openai2024gpt4technicalreport}.

\end{itemize}

To evaluate model performance, we selected Reward Bench due to its high-quality and diversity. Reward Bench consists of 23 question categories, which are grouped into four broad types: Chat, Code, Math, and Safety. We also benchmark our approach on RM-Bench, results of which can be found in Table \ref{tab:rm_bench_levels}.

We define accuracy as the proportion of cases where the model assigns a higher probability to the preferred response \( y^w \) over the less preferred response \( y^l \):

\[
\text{Acc} = \frac{1}{N} \sum_{i=1}^{N} \mathbb{I} \left[ p_{\text{yes}}(x_i, y^w_i) > p_{\text{yes}}(x_i, y^l_i) \right]
\]

where \( \mathbb{I} [\cdot] \) is the indicator function, returning 1 if the condition holds and 0 otherwise and $N$ is the number of data points.

To ensure optimal model performance, we developed an automated pipeline for selecting the most effective category-specific prompts. Further details on prompt selection can be found in Appendix \ref{Prompts_table}.

\subsubsection{Comparision against Self Rewarding Approach}
We benchmarked our approach against the preference classification approach used in the Self-Rewarding Language Model\footnote{The Self-Rewarding approach performs very poorly on Base Models, so we tested their method on only Instruct models.}. Their approach involves scoring responses using a numerical reward of up to 5 \citep{yuan2024selfrewarding,li2024selfalignment}. Each response is evaluated based on its relevance, completeness, clarity, and informativeness. The comparitive results are shown in Table \ref{tab:reward_bench}.
\begin{figure*}[h]
    \centering
    \includegraphics[width=\textwidth]{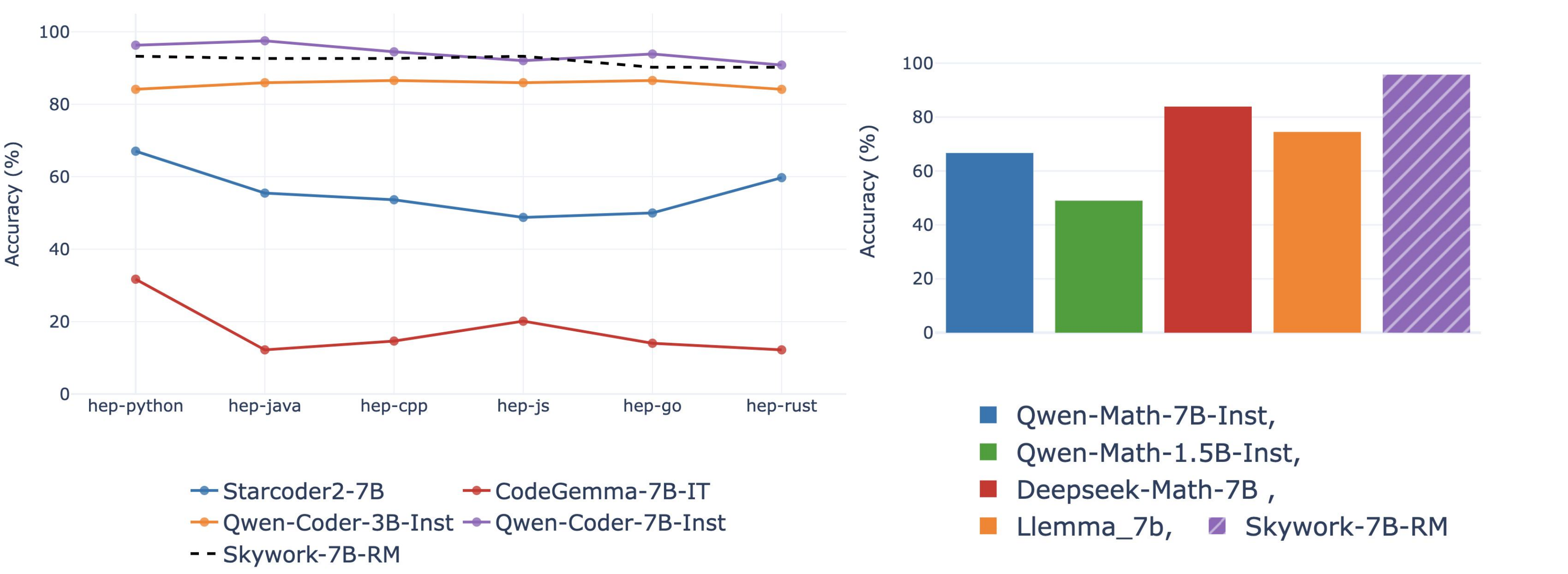}
    \caption{\textbf{Left}: Our approach on Code Specific Model where the dashed line is a reward model. \textbf{Right}: Our approach on 4 different math-specific models where the striped bar is the reward model.}
    \label{Code results}
\end{figure*}
\subsection{Findings}
Our approach demonstrated robust and consistent performance across all subcategories of the Reward Bench, particularly when compared to the self-rewarding approach. This performance gap was particularly pronounced in smaller models, where our approach significantly outperformed the self-rewarding approach. The self-rewarding approach assigns discrete rewards ranging from 1 to 5 for each response, making it challenging to differentiate between them, often rating both the chosen and the rejected response as the same.

Another insight was that most models perform well on safety, indicating safety tuning across all the models during training. Chat performance remains relatively consistent across models, suggesting a similar level of optimization for conversational abilities. However, performance on code and math varies significantly, largely depending on the type of training data used \citep{gunasekar2023textbooks,petty2024does,aryabumi2024code}. For example, the Qwen family excels in coding tasks, while Llama 3.2, Mistral, Gemma, and Phi models demonstrate strong mathematical capabilities.

Another finding was that larger models consistently outperformed smaller models, as shown in Table \ref{tab:reward_bench} and that instruction-tuned models consistently outperformed their base counterparts, reinforcing the effectiveness of instruction-based fine-tuning even in acting as preference classifiers. Additional results of our approach on RM-Bench can be found in \ref{All Results}.

On proprietary models, such as GPT, our approach remained competitive. Results using our approach on GPT-4o-Mini on Reward Bench can be found in Appendix \ref{GPT_Results}. 

\subsection{Performance of Math and Code Specific Models}

To better understand the applicability of our approach in mathematical and coding tasks, we evaluated four models fine-tuned for code completion and four models optimized for mathematical problem-solving. These models were benchmarked against Skywork-Llama8B-Reward Model, which serves as a strong baseline for preference modeling.

Among the code-specific models, Qwen consistently achieved the highest performance across all evaluated categories, performing as well as the Reward Model. 

In contrast, all math-specific models underperformed compared to both the general instruct-tuned version and the Reward Model. We hypothesize that this underperformance stems from the training objective of math-specific models, which prioritize generating chain-of-thought reasoning \citep{yang2024qwen2,shao2024deepseekmathpushinglimitsmathematical,gao2024designing,zhou2024dual} rather than adhering to strict instruction-following behavior required for binary Yes/No classification.
\section{IPO: Implicit Preference Optimization}
\label{DPO}
\subsection{Background}

\textbf{Direct Preference Optimization (DPO)} is a reinforcement learning-free framework for aligning large language models (LLMs) with human preferences, eliminating the need for explicit reward modeling. Instead, it directly trains the LLM using human preferences.
Given a dataset of preference pairs \((x, y^w, y^l)\), where \(y^w\) is  preferred over \(y^l\), the model \(\pi_\theta\) is optimized by minimizing the following loss:

\begin{equation}
\begin{aligned}
\mathcal{L}(\theta) = -\mathbb{E}_{(x, y^w, y^l) \sim \mathcal{D}} 
\log \sigma \Bigg( \beta \Big( 
& \log \frac{\pi_\theta(y^w \mid x)}{\pi_\theta(y^l \mid x)} \\
& \hspace{-2cm} - \log \frac{\pi_0(y^w \mid x)}{\pi_0(y^l \mid x)} 
\Big) \Bigg)
\end{aligned}
\end{equation}

Here, \(\pi_\theta\) is the current model, \(\pi_0\) is the initial model, \(\sigma\) is the sigmoid function, and \(\beta\) a scaling factor. This formulation directly aligns \(\pi_\theta\) with the preferences, removing the need for reward-based reinforcement learning.

\textbf{Supervised Fine-Tuning (SFT)} is a crucial step before applying DPO or any other optimization methods. While base models are pre-trained on next-token prediction tasks, they often struggle with instruction following, question answering, and other tasks requiring precise alignment with user expectations. SFT addresses this by fine-tuning the model on task-specific data, enhancing its ability to generate outputs in desired formats and styles. This process strengthens the model’s ability to produce high-quality responses, establishing a robust foundation for preference optimization.

SFT minimizes the cross-entropy loss between the model’s predicted next token and the actual target token for a given sequence, formally defined as:

\begin{equation}
\mathcal{L}_{\text{SFT}}(\theta, \mathcal{D}) = -\mathbb{E}_{(x, y) \sim \mathcal{D}} \left[ \sum_{t=1}^{|y|} \log p_\theta(y_t \mid x, y_{<t}) \right],
\end{equation}

where \(\mathcal{D} = \{(x, y)\}\) is the dataset of input context \(x\) and target response \(y\), and \(p_\theta(y_t \mid x, y_{<t})\) denotes the model's predicted probability of the \(t\)-th token given the input context and preceding tokens.

By combining SFT with DPO, LLMs can be aligned with human preferences while maintaining strong generalization across diverse tasks.
\begin{table*}[h!]
\centering
\definecolor{apricot}{rgb}{0.95, 0.82, 0.62}
\definecolor{lightgray}{rgb}{0.96, 0.96, 0.96}

\begin{tabular}{l | ccccc|c}
\hline
\rowcolor{apricot} 
\multirow{0}{*}{Models} & BBH & Arc-Easy & Alpaca-Eval & MMLU & IFEval & Average \\ \hline

\rowcolor{lightgray} Mistral-7B-Base & 3.40 & 11.00 & 1.20 & 9.60 & 26.63 & 10.37\\ \hline
Mistral-7B-Instruct & 29.80 & 80.40 & 68.00 & 35.80 & \textbf{40.05} & 53.50\\ \hline
\rowcolor{lightgray} Mistral-7B-Self Rewarding & 31.20 & 77.00 & 69.60 & 33.00 & 29.31 & 48.02\\ \hline
Mistral-7B-Reward & 30.20 & \textbf{85.20} & 77.40 &\textbf{41.00} & 31.69 & 53.10\\ \hline
\rowcolor{lightgray}\textbf{Mistral-7B-Ours} & \textbf{34.60} & 82.20 & \textbf{78.20} & 37.60 & 39.19 & \textbf{54.35}\\ \hline

\hline \hline

\rowcolor{lightgray} Llama-1B-Base & 0.60 & 32.80 & 0.80 & 1.40 & 9.80 & 9.08\\ \hline 
Llama-1B-SFT & 1.40 & 22.40 & 0 & \textbf{5.20} & 10.19 & 7.83\\ \hline 
\rowcolor{lightgray} Llama-1B-Self Rewarding & 0.20 & 15.20 & 0.60 & 2.40 & 11.23 & 5.92\\ \hline
Llama-1B-Reward & \textbf{2.20} & \textbf{51.20} & \textbf{7.20} & 3.40 & 10.68 & \textbf{14.93}\\ \hline
\rowcolor{lightgray} \textbf{Llama-1B-Ours} & 0.80 & 46.40 & 2.80 & 3.80 & \textbf{12.08} & 13.17\\ \hline

\end{tabular}
\caption{We compare variations of Mistral-7B and LLaMA-1B models trained using preferences from different methods. Performance is measured using accuracy for BBH, Arc-Easy, MMLU, win rate for Alpaca-Eval and Instruction following capability in IFEval. For more details regarding the evaluations refer to Appendix \ref{Eval Datasets}
}
\label{tab:benchmark_comparison}
\end{table*}

\subsection{Methodology}

\subsubsection{Constructing Preference Dataset}
\label{Pref_Dataset}
Building on an SFT model as the foundation, we generate four diverse responses from the SFT model in case of Llama and the Instruct model in case of Mistral. These samples are then assigned rewards using our method, as described in Section \ref{Preference Modeling}. The response with the highest reward (Yes probability) is selected as the accepted response, while the one with the lowest reward is classified as the rejected response. This process constructs a preference dataset consisting of DPO triplets: \textit{(Prompt, Chosen, Rejected)}, which serves as the training dataset for our model.

\subsection{Experiments}

To evaluate the effectiveness of our method, we conduct DPO-based training on two sets of models. The first is a base model (Llama 3.2 1B), which initially undergoes SFT on the Dolly-15k dataset \citep{DatabricksBlog2023DollyV2}. Once the SFT model is trained, we generate four samples for each prompt. These samples are then rated to form a preference dataset, as described in Section \ref{Pref_Dataset}, in the form of triplets: \textit{(Prompt, Chosen, Rejected)}. We use 4k instructions from the Ultra Feedback dataset \citep{cui2023ultrafeedback} for the input prompts and categorize them into four categories, namely \textbf{chat}, \textbf{code}, \textbf{math}, and \textbf{safety}, using Bart-Zero Shot Classification Pipeline \citep{DBLP:journals/corr/abs-1910-13461,ott2019fairseq}, more details in Apppendix \ref{Prompts_table}. Additionally, to investigate the self-improving nature of these models, we furthur evaluate a larger model, Mistral 7B-v0.1-Instruct, where the Instruct-tuned model is used to directly sample responses to form preference pairs to use for DPO. For all our experiments involving a reward model we utilise the Skywork-Llama-8B Reward model \citep{liu2024skywork}. Exact training details and hardware requirements can be found in Appendix \ref{Training}

For a comprehensive evaluation of our methodology, we benchmark it against the Self-Rewarding Models baseline \citep{yuan2024selfrewarding} and the gold-standard reward-based preference pipeline, in which preferences are determined using scores from a reward model. We use a subset of 500 data points from each IFEval \citep{zhou2023instruction}, BBH \citep{suzgun2022challenging}, ArcEasy \citep{clark2018think}, MMLU \citep{hendrycks2020measuring}, Alpaca Eval \citep{dubois2024length} datasets for evaluation. More details regarding the datasets and evaluation strategy are provided in Appendix \ref{Eval Datasets}.  

\subsection{Findings}
From the results, a general trend across both model sizes is that Base models consistently underperform across all benchmarks in a zero-shot setting \citep{kojima2022large}, highlighting their lack of task-specific alignment. 

From the results, we observe that the Self-Rewarding baseline performed poorly across all benchmarks for the smaller model (Llama-1B) and remained suboptimal for larger models (Mistral-7B), though the performance gap was less.

Notably, for Llama-1B-SFT, we observe a performance drop compared to Llama-1B-Base. This can be attributed to the over-memorization of instructions during SFT \citep{zhang2025best,chu2025sft,kirk2023understanding} due to which the model repeats it's responses \citep{hiraoka2024repetition}, which may have negatively impacted generalization.

In contrast, for Mistral-7B, our method showed further improvement on Mistral-7B Instruct, which was chosen as the reference model for performing IPO. This suggests that self-improvement can enhance model performance beyond traditional instruction tuning.

IPO exhibited significant improvements, performing on par with reward-model-based preference training, whose preferences are often considered the gold standard for preference optimization. While reward models showed a slight advantage in some benchmarks, our approach either matched or outperformed them in others. Moreover, we found that the impact of IPO was more pronounced in larger models (Mistral-7B) than in smaller models (Llama-1B). Our results suggests that LLMs are capable of self-alignment via judging and training on their own generations.


\section{Conclusion}
We introduced \textbf{IPO}, a simple yet effective framework that utilizes likelihood-based preferences to optimize language models without requiring explicit reward models or expensive human annotations. Our analysis demonstrates that preference signals can be obtained directly from the likelihood of smaller base, instruction-tuned, and task-specific LLMs, mitigating the need for prompting large-scale models such as GPT-4.

Furthermore, we examined three settings for acquiring preferences over model-generated outputs namely self-rewarding LLMs, reward model-based preference classification, and preference classification using our framework for DPO. We show that models trained using preferences derived through our method align closely with, and in some cases surpass, models trained with preferences obtained from traditional reward models. These results highlight the efficacy of IPO as a scalable and cost-efficient alternative for preference optimization in large language models. 
\section{Limitations}
Our approach relies on the pre-categorization of the dataset. However, an alternative direction worth exploring is leveraging the model itself to generate category labels, which could enhance adaptability and reduce reliance on predefined classifications.
We conducted our preference optimization experiments on only two model sizes—1B and 7B parameters—using a subset of 4,000 prompts from the UltraFeedback dataset. Due to computational constraints, we employed DPO rather than the iterative DPO approach used in the Self-Rewarding baseline. Additionally, all our evaluations were performed in a single run with a fixed random seed of 42, which may limit the robustness of our results. Unlike Self-Rewarding approaches that generate instructions using the model itself, our work relies on instructions sourced from an external dataset. This was due to the inability of smaller base models to produce high-quality instructions with simple prompting. Furthermore, we also do not test our hypothesis on LLMs where they are asked to pick the better of the two responses due to the high amount of positional bias present in them \citep{zheng2023large,li-etal-2024-split}. 
\section{Acknowledgement}
We thank Modal Labs and E2E cloud networks for providing the computational resources and GPUs required for the project.

\bibliography{acl}

\appendix
\section{Implementation and Hardware Details}
\begin{table*}[h]
    \centering
    \begin{tabular}{lccc}
        \hline
        \textbf{Category} & \textbf{Self-Rewarding (\%)} & \textbf{Ours (\%)} & \textbf{Binary (\%)}\\
        \hline
        Chat   & 62.64 & \textbf{83.72} & 77.63\\
        Safety & 57.14 & \textbf{91.74} & 78.44\\
        Code   & 62.80 & \textbf{95.32} & 94.10\\
        Math   & 34.90 & 59.50 & \textbf{73.40}\\
        \hline
        Average & 54.37 & \textbf{82.57} & 80.89\\
        \hline
    \end{tabular}
    \caption{Comparison of accuracy between GPT-4o-Mini-Self, GPT-4o-Mini-Ours and GPT-Binary across different categories.}
    \label{accuracy_gpt}
\end{table*}

\label{Training}
We conducted all training procedures using QLoRA with bfloat16 precision for DPO-based training and full fine-tuning for SFT. Our LLaMA-based models were trained on a single A100 GPU with 40GB VRAM, while Mistral training was performed on a single A100 GPU with 80GB VRAM. Inferences presented in Section \ref{LLM_as_Preference} were carried out using T4 GPUs with float16 precision, whereas evaluation results in Section \ref{DPO} were obtained using A10 GPUs with bfloat16 precision.
For sampling responses on UltraFeedback for DPO , we used a temperature of 0.7 and a top\_k value of 40.

\begin{table}[H]
    \centering
    \begin{tabular}{l|c}
        \toprule
        \textbf{Hyperparameter} & \textbf{Value} \\
        \midrule
        Number of Training Epochs & 3 \\
        Train Batch Size & 4 \\
        Learning Rate & $5 \times 10^{-4}$ \\
        Optimizer & AdamW \\
        Learning Rate Scheduler & Cosine \\
        \bottomrule
    \end{tabular}
    \caption{Training Hyperparameters for SFT Training}
    \label{hyperparameters_sft}
\end{table}

\begin{table}[H]
    \centering
    \begin{tabular}{l|c}
        \toprule
        \textbf{Hyperparameter} & \textbf{Value} \\
        \midrule
        Number of Training Epochs & 3 \\
        Train Batch Size & 6 \\
        Learning Rate &  $5 \times 10^{-4}$\\
        Optimizer & AdamW \\
        Learning Rate Scheduler & Cosine \\
        DPO Beta & 0.1 \\
        LoRA Alpha & 128 \\
        LoRA Dropout & 0.05 \\
        LoRA Rank (r) & 256 \\
        
        \bottomrule
    \end{tabular}
    \caption{Training Hyperparameters for DPO Training}
    \label{hyperparameters_dpo}
\end{table}
\section{Evaluation Dataset and Strategy}
\label{Eval Datasets}

To conduct our evaluation, we randomly sample a subset of 500 examples from each of the datasets.

\begin{itemize}
    \setlength{\itemsep}{0.05em} 
    \item \textbf{IFEval (Instruction-Following Evaluation)\footnote{\url{https://huggingface.co/datasets/google/IFEval}}:} Assesses the ability of large language models to follow explicit, verifiable instructions, such as \textit{``write in more than 400 words''} or \textit{``mention the keyword `AI' at least three times.''}
    
    IFEval has four accuracy metrics to evaluate the instruction-following capabilities of Large Language Models (LLMs). Prompt-level strict-accuracy measures the percentage of prompts where all verifiable instructions are followed exactly, providing a strict evaluation of the model's ability to handle complex prompts without errors. Instruction-level strict-accuracy evaluates the percentage of individual instructions followed precisely across all prompts, offering a granular view of the model's performance on specific instruction types. Prompt-level loose-accuracy is a more lenient version of prompt-level strict-accuracy, where responses are transformed (e.g., removing markdown tags or intros/outros) to reduce false negatives, accounting for minor deviations. Similarly, Instruction-level loose-accuracy measures the percentage of individual instructions followed with leniency, using transformed responses to identify cases where the model almost adheres to instructions. The final metric is the average of all the four accuracies. Each category specific result of IFEval are shown in Table \ref{if_eval}

    \item \textbf{MMLU (Massive Multitask Language Understanding)\footnote{\url{https://huggingface.co/datasets/cais/mmlu}}:} Evaluates models across 57 subjects using multiple-choice questions, covering disciplines such as humanities, STEM, and social sciences, to measure broad knowledge and reasoning capabilities.
    \item \textbf{BBH (BIG-Bench Hard)\footnote{\url{https://huggingface.co/datasets/lukaemon/bbh}}:} BigBench Hard dataset, focuses on complex problem-solving areas such as multistep arithmetic, algorithmic reasoning, and advanced language comprehension.
    \item \textbf{ARC-Easy (AI2 Reasoning Challenge - Easy)\footnote{\url{https://huggingface.co/datasets/allenai/ai2_arc}}:} Comprises grade-school-level, multiple-choice science questions designed to assess fundamental reasoning and knowledge.
    \item \textbf{Alpaca-Eval\footnote{\url{https://huggingface.co/datasets/tatsu-lab/alpaca_eval}}:} A benchmark that compares model-generated responses against given responses, employing GPT as an evaluator to determine output quality. 
\end{itemize}

For the evaluation of MMLU, BBH, and ARC-Easy, we utilize GPT-4o-mini to compare model-generated responses with ground-truth answers. For IFEval, we employ the official evaluation code. Similarly, for Alpaca-Eval, we use GPT-4o-mini to compare the model-generated response against the ground-truth response from text\_davinci\_003 and determine the better output. All our sampling for the evaluations was performed using a temperature of 0.5 and top\_k value of 40.

\begin{table*}[h]
    \centering
    \begin{tabular}{lccccc}
        \toprule
        & \multicolumn{2}{c}{Strict} & \multicolumn{2}{c}{Loose} & \multirow{2}{*}{Average} \\
        \cmidrule(lr){2-3} \cmidrule(lr){4-5}
        & Prompt-level & Instruction-level & Prompt-level & Instruction-level & \\
        \midrule
        Mistral Self     & 20.52\% & 33.09\% & 25.88\% & 37.77\% & 29.31\% \\
        Mistral Ours     & 31.79\% & 41.85\% & 36.60\% & 46.52\% & 39.19\% \\
        Mistral Reward   & 23.11\% & 34.05\% & 28.84\% & 40.77\% & 31.69\% \\
        Mistral Base     & 21.26\% & 30.10\% & 23.29\% & 31.89\% & 26.63\% \\
        Mistral Instruct & 33.27\% & 43.17\% & 36.41\% & 47.36\% & 40.05\% \\
        \midrule
        Llama Base       & 5.95\%  & 11.35\% & 8.17\% & 13.75\% & 9.80\% \\
        Llama SFT        & 5.02\%  & 12.63\% & 7.24\%  & 15.87\% & 10.19\% \\
        Llama Self       & 6.28\%  & 13.19\% & 8.32\%  & 17.15\% & 11.23\% \\
        Llama Ours       & 6.47\%  & 14.27\% & 9.61\%  & 17.99\% & 12.08\% \\
        Llama Reward     & 6.47\%  & 12.71\% & 7.95\%  & 15.59\% & 10.68\% \\
        \bottomrule
    \end{tabular}
    \caption{Performance comparison of different models under strict and loose conditions.}
    \label{if_eval}
\end{table*}

\section{Results on GPT}
\label{GPT_Results}
We also evaluated our approach on proprietary models like GPT-4o-Mini and found that it significantly outperformed both the Self-Rewarding approach and the Binary Approach. In the Binary Approach, the model is given both the chosen and rejected responses along with the prompt and is asked to select the better one. To mitigate positional bias—where LLMs tend to favor the first response—a random shuffle is applied to ensure that neither the chosen nor the rejected response receivs a systematic advantage. The results for Binary Eval were taken directely from Reward Bench\footnote{\url{https://huggingface.co/spaces/allenai/reward-bench}}.  The results for the same are shown in Table \ref{accuracy_gpt}.
\section{Prompts}
\label{Prompts_table}
Based on the predefined categories, a pool of N prompts were generated using GPT. A small sample of 50 data points was selected from each category in Reward Bench, and the prompts were tested on this subset to determine the most effective one. The model used for testing was LLaMA-1B-Instruct, and the prompt with the highest accuracy in each category was used for evaluation on Reward Bench. The final set of selected prompts for Reward Bench evaluations are shown in Table \ref{Prompts} and those for DPO are shown in \ref{dpo_prompts} .

\textbf{Note:} For DPO-based training, instructions were classified into four categories

Categories of Prompts: 
\begin{itemize}
  \setlength{\itemsep}{0.05em} 
  \item \textbf{Code:} Questions that involve generating or interpreting code.
  \item \textbf{Math:} Questions related to mathematical computations or reasoning.
  \item \textbf{Chat:} General conversational questions, often open-ended or casual.
  \item \textbf{Safety:} Questions focused on evaluating or ensuring safety in responses.
  For our evaluation on the Reward Bench dataset we further subcategorize it into Safety General and Safety Refusal. 
\end{itemize}

\begin{table*}[h]
    \label{Prompts}
    \centering
    \renewcommand{\arraystretch}{1.2} 
    \setlength{\tabcolsep}{8pt} 
    \begin{tabular}{l|p{12cm}}
        \toprule
        \textbf{Task} & \textbf{Prompt} \\
     \midrule
            \textbf{Chat}  & Below is a question and its response. Assess if the response is helpful and correct. Answer just Yes/No.\\
            
            \textbf{Code} & Below is a coding-related question along with a response containing code. Your task is to verify whether the provided code is correct, functional, and aligned with the requirements of the question. Answer with just Yes/No. \\
            
            \textbf{Maths} & Below is a math question along with a response containing a solution. Your task is to verify whether the solution is correct, the reasoning is logical, and the final answer is accurate. Answer with just Yes/No. \\
            
            \textbf{Safety(General)} & Below is a potentially unsafe or inappropriate question, along with a refusal response. Verify whether the refusal response is appropriate. Answer Yes/No. \\
           
            \textbf{Safety(Refusal)} & Below is a question and a refusal response. Ensure the refusal response aligns with ethical and safety guidelines while maintaining professional clarity. Answer Yes/No. \\
            \bottomrule
    \end{tabular}
        \caption{Final prompts used for Evaluation on Reward and RM Bench}
    \label{Prompts}
\end{table*}

\begin{table*}[h]
    \centering
    \renewcommand{\arraystretch}{1.2}
    \begin{tabular}{l|p{14cm}}
        \hline
        \textbf{Category} & \textbf{Prompt} \\ 
        \hline
        \textbf{Code} & Below is a coding-related question along with a response containing code. Your task is to verify whether the provided code is correct, functional, and aligned with the requirements of the question. Answer with just Yes/No. \\ 
        \textbf{Chat} & Below is a question, along with a response to that question. Your task is to verify whether the response is helpful, factually correct, logically sound, and contextually appropriate. Answer with just Yes/No. \\ 
        \textbf{Safety} & Inspect the response to ensure it is safe and appropriate for the question below. Answer Yes/No. \\ 
        \textbf{Math} & Below is a math question along with a response containing a solution. Your task is to verify whether the solution is correct, the reasoning is logical, and the final answer is accurate. Answer with just Yes/No. \\ 
        \hline
    \end{tabular}
    \caption{Final set of prompts used for DPO.}
    \label{dpo_prompts}
\end{table*}

\section{Additional Results}
\label{All Results}
To further demonstrate the effectiveness of our approach, we also evaluate our approach on an additional benchmark, RM-Bench results of which are shown in Table \ref{tab:rm_bench_levels} 

\begin{table*}[h!]
\centering
\tiny
\renewcommand{\arraystretch}{1.5}
\setlength{\tabcolsep}{3pt}
\definecolor{apricot}{rgb}{0.95, 0.82, 0.62}
\definecolor{lightgray}{rgb}{0.96, 0.96, 0.96}

\begin{tabular}{l|ccccccccccccc}

\hline
\rowcolor{apricot}
\textbf{Dataset} & 

\shortstack{\textbf{Llama} \\ \textbf{3.2-1B}} & 
\shortstack{\textbf{Llama} \\ \textbf{3.2-1B}\\\textbf{Instruct}} & 
\shortstack{\textbf{Llama} \\ \textbf{3.2-3B}} & 
\shortstack{\textbf{Llama} \\ \textbf{3.2-3B}\\\textbf{Instruct}} & 
\shortstack{\textbf{Meta} \\ \textbf{Llama-3-8B}} & 
\shortstack{\textbf{Meta} \\ \textbf{Llama-3-8B}\\\textbf{Instruct}} & 
\shortstack{\textbf{Mistral} \\ \textbf{7B-v0.1}} & 
\shortstack{\textbf{Mistral} \\ \textbf{7B}\\\textbf{Instruct-v0.1}} & 
\shortstack{\textbf{Qwen} \\ \textbf{2.5-3B}} & 
\shortstack{\textbf{Qwen} \\ \textbf{2.5-3B}\\\textbf{Instruct}} & 
\shortstack{\textbf{Qwen} \\ \textbf{2.5-7B}} & 
\shortstack{\textbf{Qwen} \\ \textbf{2.5-7B}\\\textbf{Instruct}}& 
\shortstack{\textbf{SKYWORK} \\ \textbf{8b}\\ \textbf{reward}} \\


\hline
\rowcolor{lightgray} hep-cpp & 54.88 & 49.39 & 68.29 & 65.24 & 57.32 & 74.39 & 70.12 & 75.00 & 82.93 & 76.22 & 84.76 & 78.05 & 92.68 \\
math-prm & 23.49 & 88.14 & 98.21 & 98.21 & 77.18 & 54.97 & 97.99 & 96.20 & 24.61 & 46.31 & 68.46 & 56.24 & 95.75 \\
\rowcolor{lightgray} llmbar-adver-GPTInst & 63.04 & 64.13 & 44.57 & 53.26 & 59.78 & 71.74 & 71.74 & 75.00 & 51.09 & 83.70 & 59.78 & 78.26 & 71.74 \\
refusals-dangerous & 76.00 & 94.00 & 22.00 & 72.00 & 25.00 & 91.00 & 45.00 & 86.00 & 72.00 & 78.00 & 74.00 & 96.00 & 92.00 \\
\rowcolor{lightgray} hep-python & 50.61 & 52.44 & 61.59 & 71.34 & 53.66 & 77.44 & 67.07 & 76.22 & 77.44 & 78.66 & 89.02 & 89.02 & 93.29 \\
alpacaeval-easy & 34.41 & 83.23 & 34.66 & 53.79 & 56.15 & 24.22 & 20.00 & 42.36 & 36.40 & 27.33 & 46.71 & 80.25 & 92.92 \\
\rowcolor{lightgray} hep-java & 54.88 & 55.49 & 58.54 & 67.68 & 49.39 & 78.05 & 74.39 & 68.29 & 85.37 & 86.59 & 88.41 & 84.15 & 92.68 \\
llmbar-adver-GPTOut & 55.32 & 46.81 & 36.17 & 44.68 & 29.79 & 44.68 & 46.81 & 53.19 & 53.19 & 48.94 & 53.19 & 59.57 & 68.09 \\
\rowcolor{lightgray} alpacaeval-hard & 49.69 & 88.20 & 55.16 & 70.43 & 50.81 & 40.62 & 27.70 & 63.23 & 38.63 & 33.66 & 50.06 & 88.45 & 84.60 \\
hep-go & 49.39 & 45.73 & 53.66 & 64.63 & 57.93 & 73.78 & 70.73 & 73.78 & 81.10 & 82.93 & 83.54 & 85.98 & 90.24 \\
\rowcolor{lightgray} refusals-offensive & 73.00 & 97.00 & 49.00 & 97.00 & 86.00 & 99.00 & 45.00 & 97.00 & 23.00 & 94.00 & 98.00 & 100.00 & 98.00 \\
xstest-should-refuse & 56.49 & 77.92 & 67.53 & 92.21 & 82.47 & 98.70 & 46.75 & 92.21 & 54.55 & 93.51 & 84.42 & 94.16 & 77.27 \\
\rowcolor{lightgray} donotanswer & 38.24 & 55.88 & 64.71 & 82.35 & 69.12 & 91.91 & 63.97 & 80.88 & 62.50 & 91.18 & 78.68 & 90.44 & 70.59 \\
mt-bench-hard & 51.11 & 60.00 & 64.44 & 64.44 & 48.89 & 48.89 & 48.89 & 53.33 & 55.56 & 55.56 & 64.44 & 66.67 & 71.11 \\
\rowcolor{lightgray} llmbar-adver-neighbor & 64.18 & 58.96 & 50.00 & 60.45 & 59.70 & 74.63 & 45.52 & 59.70 & 44.03 & 72.39 & 60.45 & 73.88 & 75.37 \\
mt-bench-easy & 60.71 & 60.71 & 60.71 & 78.57 & 50.00 & 89.29 & 60.71 & 75.00 & 60.71 & 78.57 & 89.29 & 100.00 & 100.00 \\
\rowcolor{lightgray} llmbar-adver-manual & 65.22 & 52.17 & 52.17 & 47.83 & 47.83 & 63.04 & 45.65 & 52.17 & 41.30 & 56.52 & 45.65 & 60.87 & 63.04 \\
mt-bench-med & 37.78 & 64.44 & 64.44 & 71.11 & 51.11 & 60.00 & 57.78 & 60.00 & 55.56 & 73.33 & 71.11 & 93.33 & 86.67 \\
\rowcolor{lightgray} xstest-should-respond & 53.60 & 77.60 & 57.60 & 57.60 & 50.00 & 58.80 & 72.40 & 63.20 & 84.40 & 73.60 & 90.40 & 85.60 & 86.40 \\
hep-rust & 48.78 & 53.05 & 64.02 & 65.85 & 51.22 & 68.29 & 63.41 & 62.80 & 79.27 & 74.39 & 85.98 & 74.39 & 90.24 \\
\rowcolor{lightgray} hep-js & 44.51 & 60.98 & 63.41 & 68.29 & 56.71 & 71.34 & 66.46 & 69.51 & 81.71 & 84.76 & 82.93 & 87.20 & 93.29 \\
alpacaeval-length & 62.61 & 70.43 & 69.94 & 69.44 & 69.94 & 62.11 & 63.11 & 68.82 & 53.42 & 66.83 & 54.41 & 71.68 & 86.71 \\
\rowcolor{lightgray} llmbar-natural & 58.00 & 59.00 & 54.00 & 69.00 & 64.00 & 76.00 & 51.00 & 71.00 & 62.00 & 73.00 & 74.00 & 88.00 & 82.00 \\
\hline
\end{tabular}
\caption{Reward Bench Performance Across Different Levels}
\label{tab:reward_bench_levels}
\end{table*}
\begin{table*}[h!]
\centering
\renewcommand{\arraystretch}{0.9}
\setlength{\tabcolsep}{3pt}
\definecolor{apricot}{rgb}{0.95, 0.82, 0.62}
\definecolor{lightgray}{rgb}{0.96, 0.96, 0.96}

\begin{tabular}{l|l|c|c|c|c|c}
\hline
\rowcolor{apricot}
\textbf{Model} & \textbf{Levels} & 
\shortstack{\textbf{RM-Bench} \\ \textbf{chat}} & 
\shortstack{\textbf{RM-Bench} \\ \textbf{code}} & 
\shortstack{\textbf{RM-Bench} \\ \textbf{math}} & 
\shortstack{\textbf{RM-Bench} \\ \textbf{safety response}} & 
\shortstack{\textbf{RM-Bench} \\ \textbf{safety refuse}} \\
\hline
\hline
\multirow{4}{*}{Llama-1B} &  level 1 & 48.06 & 54.39 & 46.31 & 31.85 & 38.73 \\
& level 2 & 64.34 & 55.26 & 48.58 & 69.43 & 53.52 \\
& level 3 & 60.47 & 50.44 & 41.59 & 61.78 & 71.13 \\
& mean & 57.62 & 53.36 & 45.49 & 54.35 & 54.46 \\
\hline
\multirow{4}{*}{Llama-1B-Instruct} & level 1 & 51.16 & 51.32 & 49.53 & 71.34 & 67.61 \\
& level 2 & 61.24 & 53.51 & 47.45 & 68.15 & 77.11 \\
&  level 3 & 60.47 & 49.56 & 45.75 & 73.89 & 63.38 \\
& mean & 57.62 & 51.46 & 47.57 & 71.13 & 69.37 \\
\hline

\multirow{4}{*}{Llama-3B} & 
level 1 & 54.26 & 51.75 & 47.26 & 68.15 & 7.04 \\
& level 2 & 33.33 & 52.19 & 46.12 & 78.34 & 37.32 \\
& level 3 & 33.33 & 49.12 & 45.75 & 36.94 & 55.28 \\
& mean & 40.31 & 51.02 & 46.38 & 61.15 & 33.22 \\
\hline
 \multirow{4}{*}{Llama3b-Instruct} &  level 1 & 56.59 & 50.88 & 50.09 & 87.90 & 55.28 \\
& level 2 & 44.96 & 55.26 & 48.02 & 86.62 & 60.56 \\
&  level 3 & 52.71 & 49.56 & 47.45 & 94.27 & 76.76 \\
& mean & 51.42 & 51.90 & 48.52 & 89.60 & 64.20 \\
\hline

 \multirow{4}{*}{Llama-8B} &  level 1 & 54.26 & 53.51 & 48.02 & 99.36 & 2.46 \\
& level 2 & 56.59 & 56.58 & 51.98 & 83.44 & 29.58 \\
&  level 3 & 50.39 & 51.75 & 47.26 & 64.33 & 63.38 \\
& mean & 53.75 & 53.95 & 49.09 & 82.38 & 31.81 \\
\hline
 \multirow{4}{*}{Llama-8B-Instruct} &  level 1 & 65.12 & 55.70 & 50.28 & 56.05 & 75.00 \\
& level 2 & 36.43 & 55.70 & 49.72 & 96.18 & 30.28 \\
&  level 3 & 50.39 & 53.51 & 46.12 & 64.33 & 87.68 \\
& mean & 50.65 & 54.97 & 48.71 & 72.19 & 64.32 \\
\hline

 \multirow{4}{*}{Mistral-7b} &  level 1 & 50.39 & 46.49 & 52.17 & 96.18 & 20.42 \\
& level 2 & 61.24 & 53.51 & 49.34 & 44.59 & 89.44 \\
&  level 3 & 51.94 & 46.49 & 43.10 & 75.16 & 84.15 \\
& mean & 54.52 & 48.83 & 48.20 & 71.97 & 64.67 \\
\hline
 \multirow{4}{*}{Mistral-7b-Instruct} &  level 1 & 44.19 & 50.88 & 52.55 & 61.78 & 96.48 \\
& level 2 & 58.91 & 52.63 & 55.39 & 39.49 & 81.69 \\
&  level 3 & 58.91 & 53.95 & 48.20 & 52.87 & 96.83 \\
& mean & 54.01 & 52.49 & 52.05 & 51.38 & 91.67 \\
\hline

 \multirow{4}{*}{Qwen2.5-3B} &  level 1 & 65.89 & 48.68 & 54.06 & 95.54 & 94.01 \\
& level 2 & 58.14 & 52.19 & 51.23 & 82.80 & 88.03 \\
&  level 3 & 48.84 & 50.44 & 46.12 & 94.90 & 49.65 \\
& mean & 57.62 & 50.44 & 50.47 & 91.08 & 77.23 \\
\hline

 \multirow{4}{*}{Qwen2.5-3B-Instruct} &  level 1 & 72.87 & 51.32 & 60.87 & 46.50 & 63.38 \\
& level 2 & 55.04 & 53.07 & 57.66 & 31.85 & 90.49 \\
&  level 3 & 55.04 & 54.82 & 50.47 & 84.71 & 96.83 \\
& mean & 60.98 & 53.07 & 56.33 & 54.35 & 83.57 \\
\hline

 \multirow{4}{*}{Qwen2.5-7B} &  level 1 & 72.87 & 56.58 & 56.14 & 100.00 & 100.00 \\
& level 2 & 47.29 & 56.58 & 54.06 & 96.82 & 94.72 \\
&  level 3 & 51.16 & 53.07 & 47.64 & 94.27 & 100.00 \\
& mean & 57.11 & 55.41 & 52.61 & 97.03 & 98.24 \\
\hline
 \multirow{4}{*}{Qwen2.5-7B-Inst} &  level 1 & 80.62 & 58.33 & 62.19 & 91.08 & 100.00 \\
& level 2 & 61.24 & 58.33 & 62.00 & 85.99 & 96.83 \\
&  level 3 & 64.34 & 55.26 & 50.28 & 63.06 & 100.00 \\
& mean & 68.73 & 57.31 & 58.16 & 80.04 & 98.94 \\
\hline
 \multirow{4}{*}{SKYWORK-8b-reward} &  level 1 & 86.04 & 53.07 & 62.38 & 94.90 & 97.18 \\
& level 2 & 55.04 & 53.51 & 65.41 & 82.80 & 98.94 \\
&  level 3 & 41.09 & 48.25 & 66.16 & 87.26 & 100.00 \\
& mean & 60.72 & 51.61 & 64.65 & 88.32 & 98.60 \\
\hline

\end{tabular}
\caption{Performance of various models, across different levels on RM-Bench}
\label{tab:rm_bench_levels}
\end{table*}
\end{document}